# HOW MUCH DOES A WORD WEIGHT? WEIGHTING WORD EMBEDDINGS FOR WORD SENSE INDUCTION

**Arefyev N.** (narefyev@cs.msu.ru)
Lomonosov Moscow State University & Samsung Moscow Research Center, Moscow, Russia

**Ermolaev P.** (permolaev@cs.msu.ru)
Lomonosov Moscow State University, Moscow, Russia

**Panchenko A.** (panchenko@informatik.uni-hamburg.de)
University of Hamburg, Hamburg, Germany

The paper describes our participation in the first shared task on word sense induction and disambiguation for the Russian language RUSSE'2018 [Panchenko et al., 2018]. For each of several dozens of ambiguous words, the participants were asked to group text fragments containing it according to the senses of this word, which were not provided beforehand, therefore the „induction" part of the task. For instance, a word "*bank*" and a set of text fragments (also known as "contexts") in which this word occurs, e.g. "*bank is a financial institution that accepts deposits*" and "*river bank is a slope beside a body of water*" were given. A participant was asked to cluster such contexts in the unknown in advance number of clusters corresponding to, in this case, the "company" and the "area" senses of the word "*bank*". The organizers proposed three evaluation datasets of varying complexity and text genres based respectively on texts of Wikipedia, Web pages, and a dictionary of the Russian language.

We present two experiments: a positive and a negative one, based respectively on clustering of contexts represented as a weighted average of word embeddings and on machine translation using two state-of-the-art production neural machine translation systems. Our team showed the second best result on two datasets and the third best result on the remaining one dataset among 18 participating teams. We managed to substantially outperform competitive state-of-the-art baselines from the previous years based on sense embeddings.

# СКОЛЬКО ВЕСИТ СЛОВО? ВЗВЕШИВАНИЕ ВЕКТОРОВ СЛОВ В ЗАДАЧЕ ИЗВЛЕЧЕНИЯ ЗНАЧЕНИЙ СЛОВ

**Арефьев Н.** (narefyev@cs.msu.ru)
Московский Государственный Университет им. М. В. Ломоносова и Московский Исследовательский Центр Самсунг, Москва, Россия

**Ермолаев П.** (permolaev@cs.msu.ru)
Московский Государственный Университет им. М. В. Ломоносова, Москва, Россия

**Панченко А.** (panchenko@informatik.uni-hamburg.de)
Университет Гамбурга, Гамбург, Германия

## 1. Introduction

Word sense induction (WSI) task aims at identification of word senses for ambiguous words in unsupervised and knowledge-free manner i.e. without using any manually compiled dictionaries or sense inventories. While a few languages, such as English, have such lexical resources of good quality and coverage the appeal of the WSI setting is the possibility to enable word sense disambiguation (WSD) for languages and domains where such resources are not available. Slavic languages still do not have lexical resources with broad coverage comparable, for instance, to English WordNet [Miller et al, 1990] which provides a comprehensive inventory of senses. The word sense induction task was thoroughly studied in the context of a few popular Western European languages, such as English, French, and German. However, for the Russian language only few word sense disambiguation experiments were performed, e.g. [Lopukhina et al, 2016] motivating the need for more research in this field.

Main research results related to word sense induction and disambiguation were effectively reported on the materials of the English language. Notably, several shared tasks performed a systematic evaluation of approaches in this field. Namely, [Agirre and Soroa, 2007] presented a SemEval task where participants were provided with contexts in English which they had to group according to word senses. The gold standard annotation used WordNet sense inventory. [Manandhar et al., 2010] presented a similar evaluation campaign, which was devoted to word sense induction of nouns and verbs. For each target word, participants were provided with a training set in order to learn the senses of that word. Then, participating systems disambiguate unseen instances (contexts) of the same words using the learned senses. [Jurgens and Klapaftis, 2013] performed an evaluation in the multi-sense labeling task. In this setup, participating systems provide a context with one or more sense labels weighted by the degree of applicability. More recently, [Alagić et al., 2018] presented an instance representation based on lexical substitutes—contextually suitable meaning-preserving replacements of words in context. [Corrêa et al., 2018] proposed a method that leverages recent findings in word embeddings research to generate context embeddings. Their word sense induction method represents a set of ambiguous words as a complex network, where edges are generated based on word embeddings similarity. [Pelevina et al., 2016] investigate another graph-based approach to word sense induction which relies on a graph clustering method applied to an ego-network of distributionally related words, which is constructed using word embeddings. [Panchenko et al., 2017] rely on a similar approach, making the induced senses interpretable using hypernymy labels, images, and definitions of senses extracted in an unsupervised way.

Some research related to word sense induction and disambiguation was also performed before for the Russian language, however not as a part of an evaluation campaign, but rather as individual contributions with often incomparable evaluation benchmarks, making it difficult to compare performance of different approaches. Loukachevitch and [Chuiko, 2007] proposed a method for all-word disambiguation task on the basis of a thesaurus. [Kobritsov et al., 2005] developed disambiguation filters to provide semantic annotation for the Russian National Corpus. The semantic annotation was based on the taxonomy of lexical and semantic facets. Lyashevskaya and [Mitrofanova, 2009] proposed a statistical word sense disambiguation models on an example of several nouns. [Lopukhin et al., 2017] evaluated several approaches: Adaptive Skip-gram, Latent Dirichlet Allocation, clustering of contexts, and clustering of synonyms. [Ustalov et al., 2017] proposed a meta-clustering algorithm for graphs designed for unsupervised acquisition of word senses and grouping them into synsets using Wiktionary and other dictionaries of synonyms.

In this paper we make a further step in this direction, improving upon the current state-of-the-art results. Our experiment is performed in a competitive setting of the first shared task on word sense induction and disambiguation RUSSE'2018 [Panchenko et al., 2018] aiming at comparing sense induction and disambiguation systems for the Russian language. Namely, we present two approaches to WSI for the Russian language. One of the approaches was the second best in two datasets and the third best in the remaining one dataset in this evaluation campaign where 18 participants submitted over a hundred of various models. Besides, our model was better that one of the state-of-the-art approaches to WSI based on AdaGram sense embeddings [Bartunov et al., 2016]; [Lopukhina et al., 2016].

The paper is structured the following way. First, in **Section 2**, we describe the positive result: an approach based on clustering of contexts represented as a weighted average of word embeddings. We used word2vec embeddings and compared different weighting schemes. This method yielded the best results. Second, in **Section 3**, we present a negative result: an alternative approach based on neural machine translation. Machine translation has improved a lot recently after the introduction of neural network based translation systems. In order to translate ambiguous words correctly a translation system should disambiguate them first, so we hoped to benefit from translating contexts of ambiguous words from Russian into English with two of the best available machine translation systems, namely Yandex Translate and Google Translate. This approach showed worse results than the first one and we analyze possible reasons. Next, in Section 4, we present a qualitative analysis of both approaches on several examples. Finally, we conclude the paper with a summary of the experiments and contributions.

## 2. Word Sense Induction via Clustering of Contexts Represented as a Weighted Average of Embeddings

This section presents **a positive result**, using an approach to word sense induction, which managed to achieve highly competitive results in the shared task.

### 2.1. Description of the Method

In this approach, the word sense induction task is formalized as a clustering task. Namely, each context of an ambiguous word is represented as a weighted average of all of its words' embeddings with carefully selected weights. Finally, all contexts of the ambiguous word represented as vectors in the same space are clustered. Different clusters are interpreted as different senses of the ambiguous word. The three steps of the method are described below.

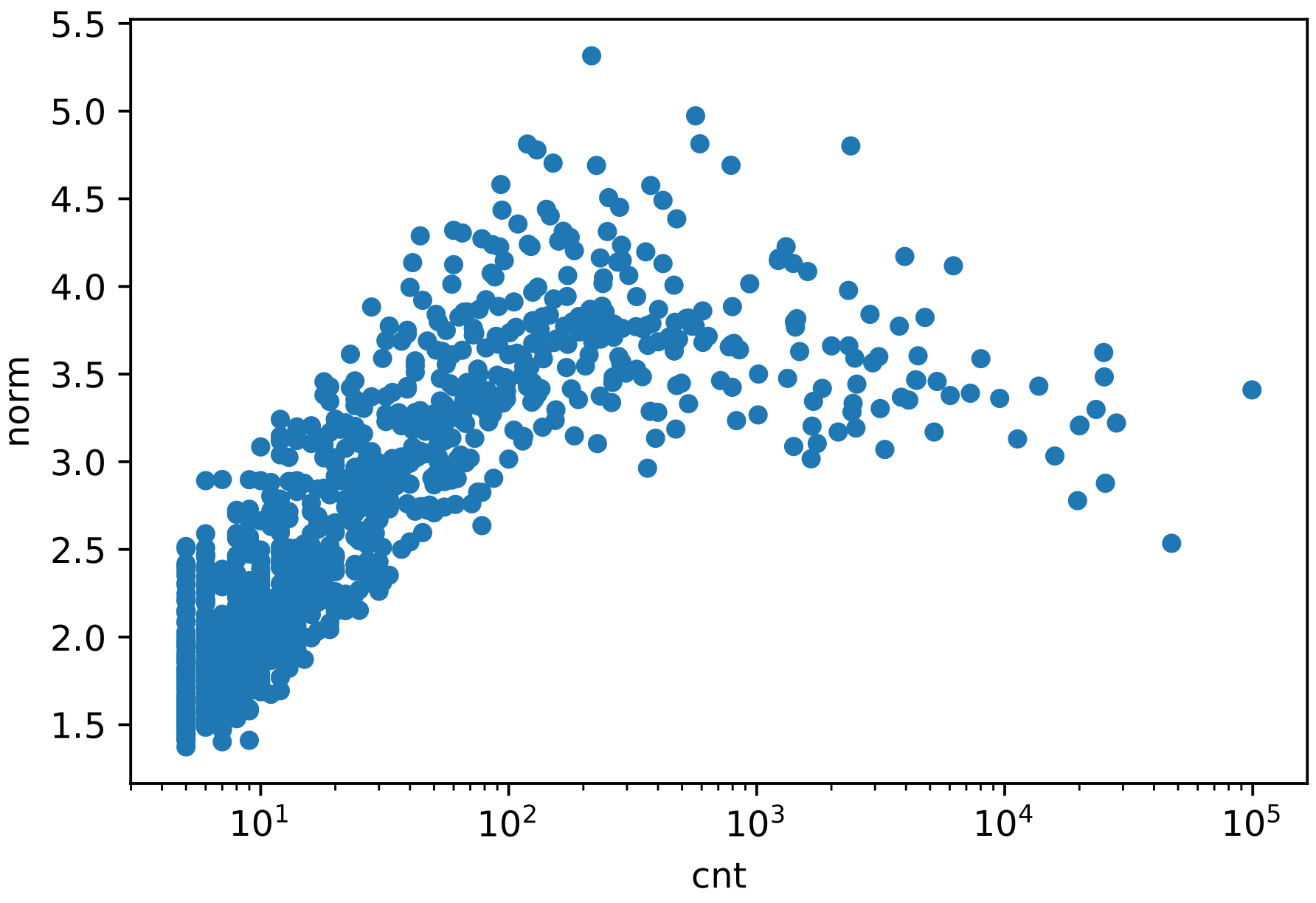

**Figure 1.** L2 norm of skip-gram word vectors as a function of word frequency: the higher the word frequency the larger the norm of this word embedding

### 2.1.1. Learning Word Embeddings from a Large Unlabeled Corpus

The train and the test datasets of the shared task are fairly small making the representations learned only from them perform poorly. For this reason, the information from the background collections is exploited in the form of word embeddings. Namely, we trained several word2vec models [Mikolov et al., 2013], both CBOW and Skip-gram, using different corpora: the corpus of books in Russian with 150 Gb of plain text extracted from lib.rus.ec library in [Arefyev et.al., 2015], [Panchenko et al. 2016] and the Russian Wikipedia containing about 3 Gb of plain text. Also, we experimented with various hyperparameters including window size, minimum word frequency, and corpora preprocessing type. During preliminary experiments on the training data, for active-dict and bts-rnc datasets, we have chosen the Skip-gram model with window size 10 and word embeddings of size 200 trained for 3 epochs on Librusec and containing only words with at least 5 occurrences (resulting in 3 million words vocabulary). For wiki-wiki dataset, we have selected a model with similar hyperparameters trained on the Russian Wikipedia which performed better for this dataset.

It is common to normalize word2vec embeddings to unit length when they are used for word similarity estimation. However, according to our experiments on the train datasets, unnormalized embeddings provided better results and we decided to perform no normalization. To explain this we plot the dependency between word frequency and corresponding embedding length for a random sample of 1 thousand words (cf. **Figure 1**). One can see a strong positive correlation with shorter embeddings for rare words and longer embedding for more frequent ones. This dependency does not hold only for the most frequent words which are few. Shorter embeddings affect weighted average less compensating for lower quality of embeddings for rare words which can explain better performance of unnormalized words vectors.

### 2.1.2. Representing Contexts in a Vector Space

To represent each example as a vector we calculated a weighted average of word-2vec embeddings for all context words, i.e. all words excluding the target (polysemous) word and all of its occurrences in all grammatical forms (this helped reducing noise introduced by different senses of the target word mixed in its embedding). An appropriate selection of the weights turned out to be critical for the overall performance as we show later. We experimented with unweighted average, tf-idf weights and chi-square statistic values (for shortness we use the term “chi2” from now on). Tf-idf weights were estimated on Russian Wikipedia. Tf-idf weights help lowering the importance of the most frequent words which are likely non-informative and should not affect the context vector much. However they do not help to differentiate between less frequent context words which are related to the target and the ones which appeared with the target by chance. To give the former more weight we used chi2 measuring independence between context words and target words. If some context word occurs frequently but only with a single target word it is natural to consider it a good feature for discriminating between this target word’s senses. Chi2 weights are higher for those context words which appear mostly with a particular target word compared to those context words which appear with different target words uniformly. For instance, high chi2 weights were given to context words like “*страхования*” (*insurance*) appearing mostly with the target word “*полис*” (*policy / city-state*), “*леса*” (*forest*) appearing mostly with "*опушка*" (*woodside / trimming*) and so on. Our best results were achieved by raising to the properly selected powers and then multiplying tf-idf and chi2 weights. We experimented with normalization of the weight vector and the resulting weighted average vector and found that normalizing both of them using L2 norm works best.

#### 2.1.3. Clustering of Word Context Vectors

Finally, for each target word separately we clustered all of its examples' context vectors. In the preliminary experiments, we tried different clustering algorithms including DBSCAN and its extensions HDBSCAN and OPTICS, Affinity Propagation, Spectral clustering, Agglomerative clustering as implemented in scikit-learn[1], pyclustering[2] and hdbscan[3] software libraries. We used only Affinity Propagation [Frey & Dueck, 2007] and Agglomerative clustering algorithms in the final experiments as they have shown the most promising results on the train datasets.

### 2.2. Experiments and Discussion of the Results

#### 2.2.1. Optimization of Hyperparameters on the Train Sets

In our experiments, we tried different word embedding models, clustering algorithms and word weighting methods for the context words. For each word in a context we look up this word's embedding and multiply it by a weight. This weight is a product of tf-idf and chi2 weights raised to different powers which were selected individually for each dataset on the corresponding train set from 6 values between 0 and 2.5 (totally 36 combinations of powers).

Despite the variable number of senses per word in the datasets, Agglomerative clustering with a fixed number of clusters showed the best results for active-dict and bts-rnc. However, for wiki-wiki, the best performance was achieved using Affinity Propagation which determines the number of clusters automatically resulting in a different number of clusters for different words. The following hyperparameters and their values were evaluated on the train set of each dataset individually and the best performing values were used for the test set.

- **Agglomerative clustering** is a bottom-up hierarchical clustering approach, where each data point is placed in its own cluster first, and the the most similar clusters are merged as one moves up the hierarchy[4]. The method has three meta-parameters:
    - *number of clusters* (from 1 to 14, 2 by default)
    - *distance between points* (euclidean, L1, L2, manhattan, cosine; all of our best result use euclidean distance so we do not mention it further)
    - *linkage* criterion defines the distance between clusters (average, ward, complete, ward by default)

1 http://scikit-learn.org

2 https://github.com/annoviko/pyclustering

3 https://github.com/scikit-learn-contrib/hdbscan

4 https://en.wikipedia.org/wiki/Hierarchical_clustering

- **Affinity Propagation clustering** is based on the message passing metaphor and finds “exemplar” members representative of clusters. The algorithm chooses the number of clusters based on the provided input data. However, the algorithm has two meta-parameters[5]:
  - *damping* controls the probability for a point to change it’s cluster (between 0.5 and 1, 0.5 by default);
  - *preference* affects the probability of creating a new cluster and hence the number of clusters (chosen between −20 and 5).

**Table 1.** Hyperparameters selected on the train sets when word2vec model trained on lib.rus.ec is used. The models in bold were ranked 2nd best in the final ranking (cf. **Table 3**)

| Dataset | Clustering Algorithm | Clustering Hyperparameters | Word Weights | Train ARI |
|---|---|---|---|---|
| wiki-wiki | Agglomerative Clustering | n_clusters=2 linkage=ward | $tfidf^{1.5}$ * chi2 | 0.8057 |
| | Affinity Propagation | damping=0.5 preference=−6.8 | $tfidf^{1.5}$ * $chi2^{0.5}$ | 0.8148 |
| bts-rnc | **Agglomerative Clustering** | **n_clusters=10 linkage=average** | **$tfidf^{1.5}$ * $chi2^{0.5}$** | 0.2633 |
| | Affinity Propagation | damping=0.9 preference=−2.9 | $tfidf^{2}$ * $chi2^{2}$ | 0.1448 |
| active-dict | **Agglomerative Clustering** | **n_clusters=3 linkage=ward** | **$tfidf^{1.5}$ * $chi2^{0.5}$** | 0.2535 |
| | Affinity Propagation | damping=0.5 preference=−1.0 | tfidf * chi2 | 0.2414 |

The results of hyperparameter selection for each of the best clustering algorithms when the Skip-gram word2vec model trained on lib.rus.ec is used are presented in Table 1. The second best result in the final ranking, according to the private test ARI score, was obtained on two of the datasets using these hyperparameters. However, on the wiki-wiki dataset simple average of word2vec vectors trained on lib.rus.ec without weights performed better according to the public test score so it was used for the final submission (see Table 3 for the final ranking). It is interesting that the best results in Table 1 were achieved with the same weighting scheme for all datasets (powers 1.5 and 0.5 for tf-idf and chi2 respectively) but with different clustering algorithms. Namely, Affinity Propagation was better than Agglomerative Clustering on wiki-wiki but worse on the other two datasets. This fact and also much better ARI scores on wiki-wiki can be explained by more coarse-grained senses in this dataset and hence better separability of context vectors which is necessary for algorithms like Affinity Propagation to select the correct number of clusters automatically.

[5] https://en.wikipedia.org/wiki/Affinity_propagation

Next, we investigate how much a properly selected weighting scheme affects the results. Performance of several variations of our method with different word weighting schemes is represented in Table 2. For each clustering algorithm and a weighting scheme the best weight powers and clustering algorithm hyperparameters were selected on the train sets. Tf-idf weights always outperform chi2 weights and their combination is always the best weighting scheme.

Figure 2 shows how ARI depends on the number of clusters and linkage for agglomerative clustering. Ward linkage with 2–3 clusters shows uniformly good results on all datasets. However for bts-rnc average linkage with 10 clusters performs little better.

**Table 2.** Impact of the weighting scheme and the clustering algorithm on performance

| Method | Word Weights | wiki-wiki | bts-rnc | active-dict |
|---|---|---|---|---|
| Agglomerative Clustering | the best power score (cf. Table 1) | 0.8057 | 0.2633 | 0.2535 |
| | only tf-idf | 0.7882 | 0.2618 | 0.2451 |
| | only chi-squared score | 0.7776 | 0.2355 | 0.2157 |
| | no weights | 0.6565 | 0.2237 | 0.2147 |
| Affinity Propagation | the best power score (cf. Table 1) | 0.8148 | 0.1448 | 0.2414 |
| | only tf-idf | 0.7623 | 0.1406 | 0.2335 |
| | only chi-squared score | 0.7525 | 0.1371 | 0.1908 |
| | no weights | 0.7866 | 0.1108 | 0.1950 |

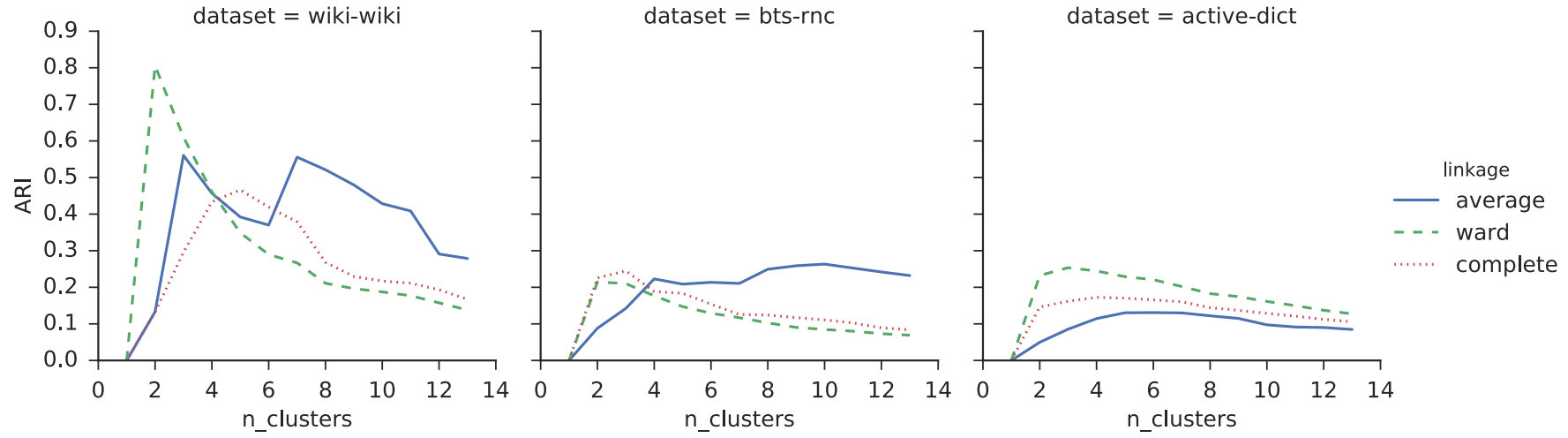

**Figure 2.** Performance of word sense induction on or the train sets depending on the number of clusters and the linkage for agglomerative clustering

Figure 3 shows a correlation between weights powers and ARI scores. For each dataset we plot a heatmap for the clustering algorithm which achieved the best score on that dataset with carefully selected hyperparameters and with the default ones. The figure shows that simple multiplication of tf-idf and chi2 weights does not work significantly better than just using tf-idf weights. It is important to properly adjust tf-idf and chi2 weights bringing them into the same scale to obtain really good improvement.

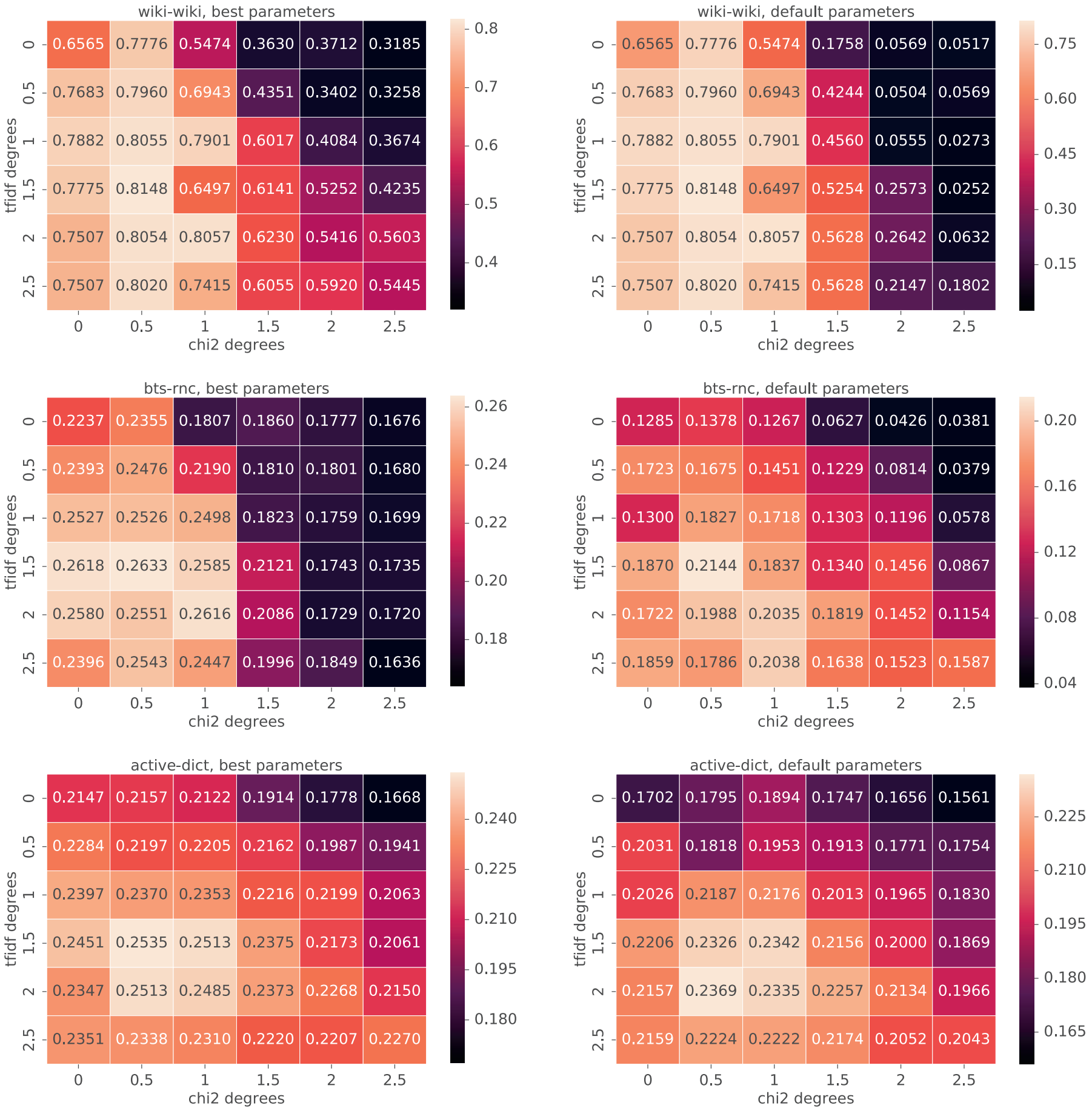

**Figure 3.** ARI on the train sets depending on powers of tf-idf and chi2 weights for the best clustering algorithm (cf. **Table 1**) with default hyperparameters and selected hyperparameters

### 2.2.2. Submitted Results for the Best Models Identified on the Train Set

Finally, Table 3 shows our best-submitted results. As one can observe, we obtain very competitive results scoring second on two datasets (bts-rnc and active-dict) and third on the wiki-wiki dataset. Furthermore, for the wiki-wiki dataset, where we ranked third, a difference with the second best participant is relatively small, as compared to the difference between the first and the second participants. We conclude that the developed methods are highly competitive with the state-of-the-art methods for word sense induction for the Russian language.

**Table 3.** The best results of our experiments, which were submitted to the RUSSE shared task. The final rank among 18 other participants is indicated in the round brackets

| Dataset | Word2vec | Weights | Clustering | Train ARI | Test ARI (public) | Test ARI (private) |
|---|---|---|---|---|---|---|
| wiki-wiki | Wikipedia | — | Affinity Propagation | 0.7577 | 1.0000 | 0.6586 (3) |
| | The second best submission in the shared task (akutuzov) | | | — | 0.9823 | 0.7096 (2) |
| | The best submission in the shared task (jamsic) | | | — | 1.0000 | 0.9625 (1) |
| bts-rnc | lib.rus.ec | $tfidf^{1.5}*chi2^{0.5}$ | Agglomerative | 0.2633 | 0.2812 | 0.2818 (2) |
| | The best submission in the shared task (jamsic) | | | — | 0.3508 | 0.3384 (1) |
| active-dict | lib.rus.ec | $tfidf^{1.5}*chi2^{0.5}$ | Agglomerative | 0.2535 | 0.2361 | 0.2270 (2) |
| | The best submission in the shared task (jamsic) | | | — | 0.2643 | 0.2477 (1) |

## 3. Using Neural Machine Translation for Word Sense Induction and Disambiguation

This section presents **a negative result**. We exploited two state-of-the-art neural machine translation systems hoping they are good enough at word sense disambiguation since it is a necessary prerequisite for good translation. Despite the fact that this approach failed, we describe it here in order to share knowledge we obtained during this study about weaknesses of currently available machine translation systems and their application to word sense induction.

### 3.1. Description of the Method

Different senses of a polysemous word in Russian are often translated using different words in English. If a translation system is good enough to produce correct translations, we could find an English word it used to translate a polysemous Russian word and utilize it as a sense identifier. Then we simply group together all examples with the same sense identifier. To implement this approach one needs to decide how to find the word in the translated text corresponding to the polysemous word in the source text. Also, there are often several occurrences of the polysemous word probably in different forms in the source text. Next, we describe how we deal with these issues.

### 3.2. Discussion of the Results

In the preliminary experiment we compared two machine translation systems available online, namely Google Translate[6] and Yandex Translate[7]. All examples of several words from active-dict and bts-rnc were translated using each system, the annotators were instructed to select the translation of the first occurrence of the target word using system's alignment (both systems highlight phrases in the source text corresponding to the selected phrase in the target text) and use it as a sense identifier for the target word. We did not try automatic alignment methods like fast_align [Dyer et.al., 2013] due to the lack of time. Hence the reported results are a kind of upper bound for machine translation approach to word sense induction.

Figures 4 and 5 show the results of the machine translation based method compared to word2vec weighted average clustering method for several words from active-dict and bts-rnc train sets. Notice that only subset of words from the train sets was annotated, so this average ARI could not be compared to the average ARI on the whole train set reported in 3.1. The results of MT systems were relatively bad compared even to the non-tuned approach from 3.1 (word2vec average with tf-idf weights followed by agglomerative clustering with 2 clusters) and the tuned version with both tf-idf and chi2 weights. To improve the results we tried normalizing translations (yandex-normalized, google-normalized in the figures) using the Porter stemmer from NLTK[8]. This improved average ARI slightly on bts-rnc and worsened it on the active-dict.

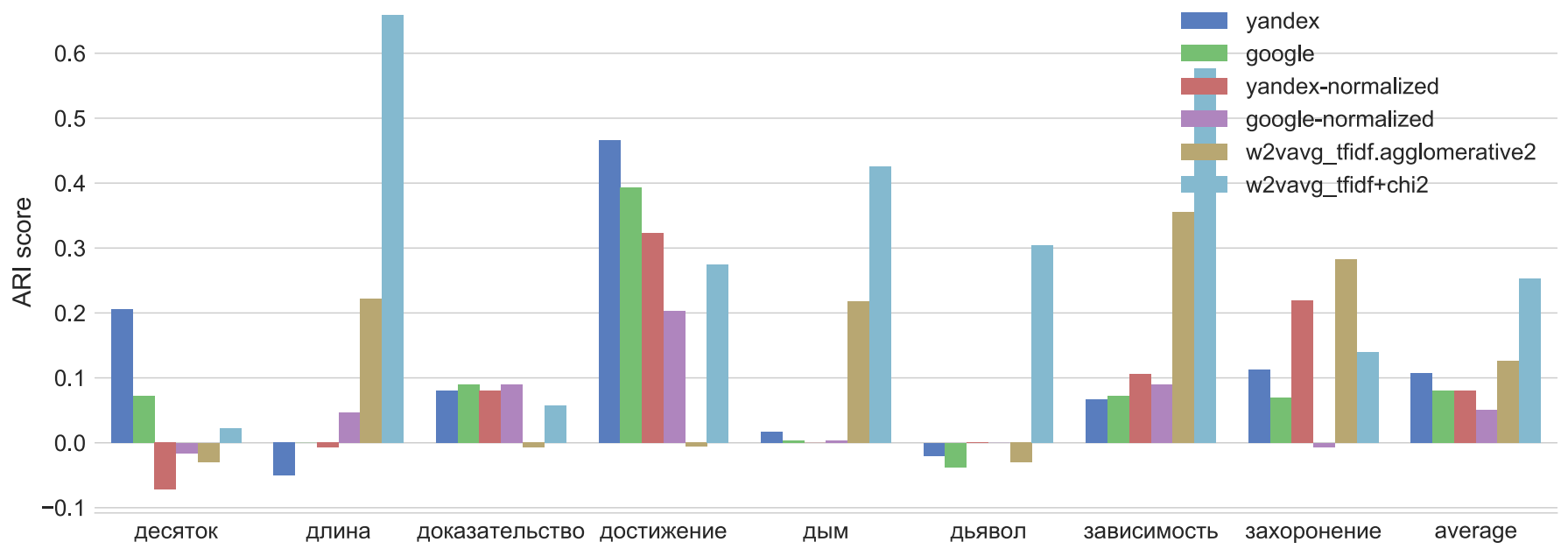

**Figure 4.** Comparison of machine translation and weighted word2vec average methods on the train set. ARI for several words from the active-dict train set and their average is presented

[6] http://translate.google.com, accessed on 21.12.2017–24.12.2017.

[7] http://translate.yandex.ru, accessed on 21.12.2017–24.12.2017.

[8] http://www.nltk.org

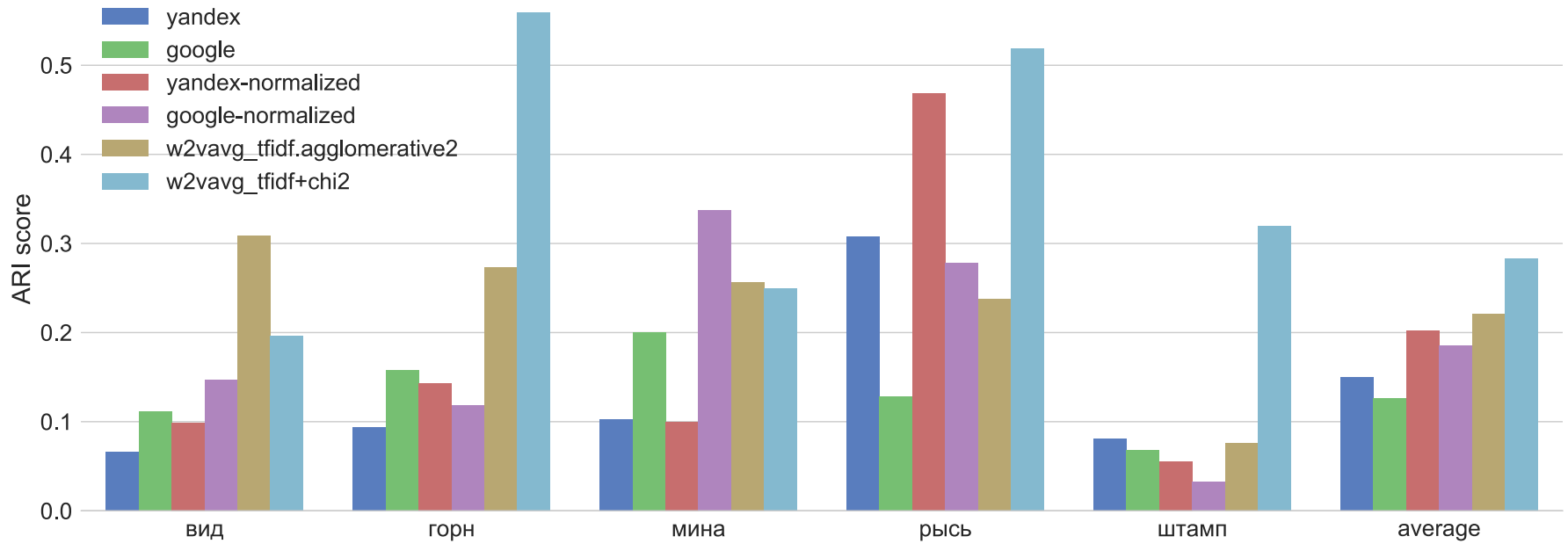

**Figure 5.** Comparison of machine translation and weighted word2vec average methods on the train set. ARI for several words from the bts-rnc train set and their average is presented

**Table 4.** A context from the wiki-wiki dataset translated by neural machine translation systems. All occurrences of the word “банка” in the source text should be translated as “jar”.

| | |
|---|---|
| source | трехлитровая **банка** во времена СССР такие банки были популярны для маринованных овощей , овощных и фруктовых соков и так далее . популярность трехлитровых банок объясняется тем , что это самая объемная банка из массово доступных , и это удобно при большом объеме заготовок . в наши дни стеклянные банки продолжают использоваться в быту для домашнего консервирования . подготовка абсолютно целых ( без трещин и сколов ) стеклянных банок подразумевает не только тщательное мытье внутри и снаружи слабым |
| Yandex Translate | three-liter **jar** in Soviet times, such banks were popular for pickled vegetables , vegetable and fruit juices and so on . the popularity of three-liter cans is explained by the fact that this is the largest Bank of massively available , and it is convenient with a large volume of blanks . nowadays, glass jars continue to be used in everyday life for home canning . training brand whole ( without cracking and chipping ) glass jars involves not only a thorough wash inside and outside of the weak |
| Google Translate | three-liter **bank** during the Soviet Union, such banks were popular for pickled vegetables, vegetable and fruit juices and so on. the popularity of three-liter cans is explained by the fact that this is the most voluminous bank of mass available, and this is convenient for a large volume of blanks. Today glass banks continue to be used in everyday life for home canning. the preparation of absolutely whole (without cracks and chips) glass jars implies not only thorough washing inside and out with the weak |

After preliminary experiments on active-dict and bts-rnc we hypothesized that the performance of MT may be better for the wiki-wiki dataset which mostly contains words with coarse-grained senses, so the next experiment was performed on the wiki-wiki test set and the results were submitted to the leaderboard. We noticed that Yandex Translate produced less fluent translations but gave better ARI, so it was chosen for this experiment. Also this time we instructed our annotators to select translations of all occurrences of the target word. To reduce errors each example was annotated by two annotators and the differences (which were very few) were eliminated by the third one. All translation were normalized and the most frequent translation for each example was used as its sense identifier. The resulting submission received 0.8125/0.3957 public/private ARI ranking 4th/12th correspondingly.

To explain the poor performance of MT-based approach we performed error analysis and noticed that translation systems are very inconsistent in their translation of polysemous words. A different occurrence of these words in a single text is often translated differently (cf. Table 4 for an example). One explanation of this inconsistency is that MT systems are trained on pairs of sentences, not pairs of texts, since currently available machine learning algorithms have problems dealing with long sequences. For instance, Yandex Translate splits input text into sentences and translate each sentence in isolation, hence it cannot take into account context from neighboring sentences[9]. We however, cannot be sure that Google Translate also performs sentence based translation. However, its performance for WSI is no better than that of Yandex.

Despite the great improvement in machine translation quality after switching from phrase based to neural based systems there is still a very large gap between machine and human translation quality and it is very unlikely to disappear in the next few years. For instance, English→French machine translation quality measured by BLEU score on news-test-2014 dataset improved from 37 to 41 points over the last 3 years while the professional human translator quality is estimated as 50 points[10]. Word sense disambiguation is probably one of the biggest challenges which should be solved for machine translation systems in order to outperform humans.

## 4. Error Analysis

In this section, we analyze the errors made by both approaches described in this paper. Consider the words “*горн*”, “*рысь*”, and “*штамп*” from the bts-rnc dataset and “*дым*”, “*зависимость*” from the active-dict dataset. Confusion matrices for these words are presented in Figure 6, their rows correspond to the true sense identifiers and our interpretation of them (in round brackets) and their columns represent either clusters built by the weighted word2vec average method or translations of the target word by the machine translation based method (the translations were stemmed resulting in absence of several few letters). The cell values are the number of examples (contexts).

[9] This information was provided in personal communication from a Yandex Translate developer.

[10] https://www.eff.org/ai/metrics#Translation

For the word “*штамп*” we could not distinguish the second sense from the first one. The other three senses (stamp, press die and cliche) are distinguished by the weighted average method in majority of examples while the machine translation uses the word “*stamp*” instead of “*cliche*” for the third sense most of the time. For the word “*дым*” (smoke, mist or household in ancient Russia) the machine translation method uses only the word “smoke” while the other method doesn’t distinguish the first and the second senses but correctly separates the household sense. The three senses of the word “*зависимость*” (mathematical correlation, political or economic dependence and addiction) can really be translated with the same word “dependence” so the base hypotheses behind machine translation approach to WSI doesn’t hold in this case and the approach fails. Conversely, the weighted average method makes very few mistakes for this word.

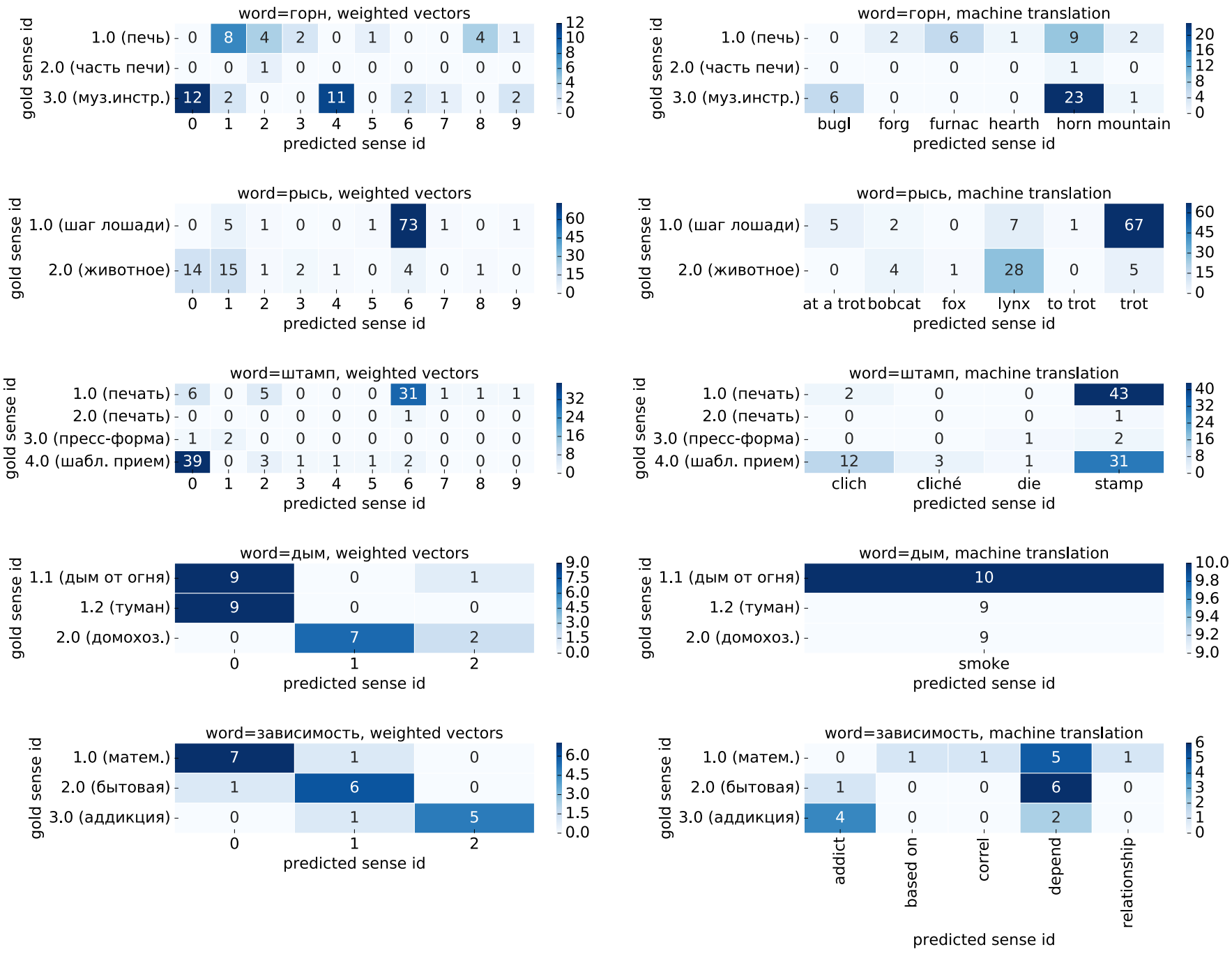

**Figure 6.** Confusion matrices for weighted word2vec average and machine translation methods

From these examples we can make a conclusion that the machine translation systems sometimes don’t work for WSI simply because they correctly use the same word to translate both senses, but more often because they don’t translate ambiguous words correctly. Although the weighted word2vec average method shows better results, sometimes it also confuses quite distant senses (stamp and cliche sense of the word “*штамп*”, for instance). Also it is unsatisfactory that on two out of three datasets simple clustering algorithms with the same number of clusters for all words work better than more advanced algorithms which select the number of clusters for each word individually. We plan to investigate the reasons of these problems and to search for a solution in the future work.

## 5. Conclusion

The paper describes two experiments on word sense induction and disambiguation for the Russian language: a positive and a negative one in terms of their results. The first (successful) one is based on clustering of contexts represented using a weighted average of word embeddings. The second (unsuccessful) is based on the state-of-the-art neural machine translation systems: the translations of ambiguous words into a different language are used as sense labels. Results of the evaluation campaign RUSSE'2018 show that the first approach yields very competitive results, compared to other 18 participating teams, ranking second on two datasets and third on the rest one. Besides, our method substantially outperforms competitive state-of-the-art baselines based on the AdaGram word sense embeddings. Interestingly, despite the expectations, the second approach based on a sophisticated production machine translation systems yielded non-competitive performance.

## 6. Acknowledgements

Alexander Panchenko was supported by Deutsche Forschungsgemeinschaft (DFG) under the projects "JOIN-T" and "ACQuA".